\documentclass[runningheads]{llncs}
\usepackage{amsmath,amsfonts,amssymb}
\usepackage{graphicx}
\usepackage{xcolor}

\begin{document}
\title{QuasiNet: a neural network with\\trainable product layers}
\titlerunning{QuasiNet: a neural network with trainable product layers}

\author{Krist\'ina Malinovsk\'a\orcidID{0000-0001-7638-028X} \and
Slavomir Holenda \and\\
L\kern-2.5pt'\kern-0.5ptudov\'it Malinovsk\'y\orcidID{0009-0000-1911-8453}}

\institute{
Faculty of Mathematics, Physics and Informatics\\
Comenius University in Bratislava, Slovakia\\
\email{kristina.malinovska@fmph.uniba.sk}\\
\texttt{http://cogsci.fmph.uniba.sk/cnc/}\\
}
\maketitle
\begin{abstract}
Classical neural networks achieve only limited convergence in hard problems such as XOR or parity when the number of hidden neurons is small. With the motivation to improve the success rate of neural networks in these problems, we propose a new neural network model inspired by existing neural network models with so called product neurons and a learning rule derived from classical error backpropagation, which elegantly solves the problem of mutually exclusive situations. Unlike existing product neurons, which have weights that are preset and not adaptable, our product layers of neurons also do learn. We tested the model and compared its success rate to a classical multilayer perceptron in the aforementioned problems as well as in other hard problems such as the two spirals. Our results indicate that our model is clearly more successful than the classical MLP and has the potential to be used in many tasks and applications.
\end{abstract}

\section{Introduction}

Multiplication of neuronal activations has been theoretically studied, but it is used relatively rarely in practice. In the simplest case, a neuron similar to a classical perceptron computes a weighted sum of the terms of a higher-order polynomial, and only the summation weights are trained via error backpropagation. Since the number of possible terms grows exponentially with increasing polynomial degree, the composition of the polynomial can be constrained in various ways. In the more complex case of the so-called product neurons, the network would learn the exponents of the individual components of each term.
However, this leads to computations in the domain of complex numbers in the general case if a negative number is raised to a fraction. 
As a result, multiplication is only used in particular cases as an optimization technique at pre-selected exact locations in the model. Most of the time, the network itself does not learn to use multiplication. Despite these obstacles, both theoretical and practical results argue in favor of using multiplication as a tool to increase the computational power of neural networks.

In this paper, we propose a novel neural network model QuasiNet that has layers of product neurons as well as layers of classical summation neurons. The novelty of our approach is that we define a new way of applying weights at the product layer, which we call quasi-exponentiation. Unlike other models, our method allows trainable product neuron weights using a gradient descent method without the need for computations in the domain of complex numbers. Thus, the network is able to partially learn the polynomial composition, which increases the generality of the model. The multiplication layer is designed in such a way that it can be plugged into the architecture as arbitrarily and easily as a classical Sigmoidal summation layer. The results of our experiments show the incomparably higher efficiency of our model in XOR and parity problems compared to a classical two-layer perceptron with the same architecture and parameters, as well as in other hard problems such as the separation of two-spirals. 

\section{Related work}

Multiplication has been used in neural models since the late 1950s, and in the 1960s it was popular to use higher-order Sigmoidal neurons for pattern recognition \cite{nilsson1965}. These were classical Sigmoidal neurons, but the input vector was augmented with terms generated by multiplying and exponentiating the inputs, and these terms were given their own weights. This approach is a natural extension of the classical perceptron, but in the general case leads to an exponential increase in dimensionality, with the upper bound on the degree of the polynomial being mainly a designer's choice. In such networks, back propagation of error can be easily applied, and neurons of this type are also known as \emph{sigma-pi} neurons \cite{rumelhart1986general}, since they perform a weighted sum of the products of the inputs.

The solution to the dimensionality problem was a family of models in which the number of polynomial terms was restricted to some pre-selected ones. For example, Giles et al.~\cite{giles1987} and Spirkovska and Reid~\cite{spirkovska1994} show that classical perceptron networks can be significantly outperformed in both learning speed and generalization capability in this way. However, this approach requires some knowledge of the problem to be embedded in the model by its designer. Constructive models have been a response to this. Their essence was to learn in different ways which polynomial terms were needed, starting with a minimal model and gradually adding higher order terms and evaluating their effectiveness \cite{redding1993,heywood1995}.

A generalization of the above principle is the product neuron model \cite{durbin1989}, which assumes trainable real weights acting as exponents of the inputs. The authors have solved the problem with the input to the domain of complex numbers by using only the real part of the number. For logical inputs, such a neuron becomes a de facto summation neuron which has cosine as an activation function, leading to further theoretical problems \cite{anthony1999}.

An alternative approach was suggested by Ghosh and colleagues \cite{ghosh1992}. Their model first linearly combines the inputs and then directly multiplies this linear combination without using an activation function. The nonlinear activation function is applied only at the output. The only primes of the model are the linear weights, and the weights on the multiplication layer are fixed to one. The result can be analyzed as a polynomial. Hence the product of the sums stands out in the formula, it is said to be a pi-sigma network. Since the weights on the multiplication layer are fixed and only linear weights are learned, the learning algorithm is based on the gradient method.

The authors tested the success of the network on many common tasks such as feature approximation, data classification, feature recognition with translation and rotation invariance, and also verified the network's ability to learn logical functions of parity, symmetry and negation. The achieved success rate is an order of magnitude better in the case of parity compared to the classical multilayer perceptron and comparable to other models with higher order neurons. 

From our point of view, the drawback is that \cite{ghosh1992} fixes the multiplicative weights and does not even try to find a way to learn them. Moreover, the absence of an activation function after the linear transformation reduces the pi-sigma network to a polynomial model with forward-selected terms. \cite{shin1995} propose a solution with a constructive model, similar to the constructive pi-sigma models above.

Analogs of multiplication occur in biological neural networks in various forms. One possibility are the so called dendritic clusters \cite{mel1994} found in the dendritic trees, which can also divide, or could represent an operation similar to the sigma-pi neurons \cite{mel1989}. The ability of neurons to multiply inputs has been found in the auditory \cite{suga1990} and visual \cite{andersen1985} cortices. Some authors argue that the physical proximity of synapses leads to multiplicative behavior, whereas distant synapses function in a summation mode \cite{bugmann1991}.

Schmitt~\cite{schmitt2002} analyzes different types of neural networks with multiplicative and sigmoidal neurons and derives general mathematical constraints for each type of network in terms of the Vapnik-Chervonenkis dimension. It concludes that multiplication in neural networks is a suitable choice for increasing the degree of nonlinear interaction and computational power, even for high-order neurons. The paper also provides a qualitative overview of neural models with multiplication, which is still quite relevant despite the current boom of deep networks with a large number of parameters and the use of brute force graphics cards.

In the context of deep neural networks, sum-product multiplication has been studied in sum-product networks \cite{poon2011,delalleau2011}. Some new models use multiplication of vectors and activation matrices \d{by elements} at selected locations of the model \cite{erhan2009,zhu2018,diba2017,schenck2018}. Articles of varying quality can be found on the blogosphere that describe specific applications of multiplication in deep neural networks and show its benefits in concrete examples. Thus, this is still an active and promising research direction.
 
\section{QuasiNet}

\begin{figure*}[ht]
	\centering
	\includegraphics[width=12cm]{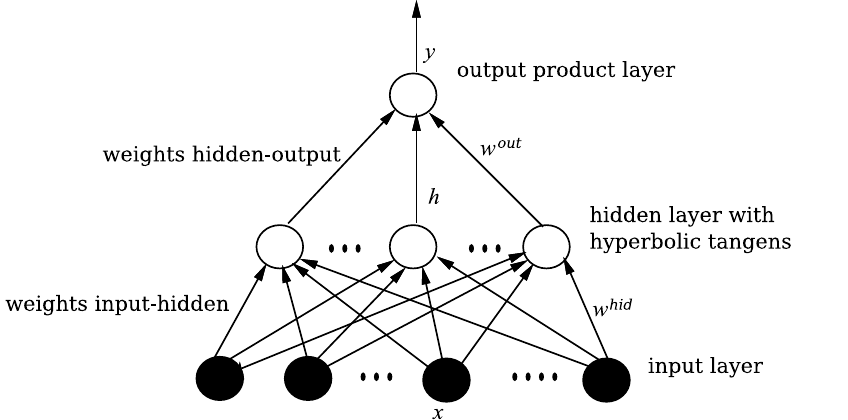}
	\caption{The schematic depiction of our model adapted from \cite{ghosh1992}.}
	\label{fig:ourmodel}
\end{figure*}

We propose a novel neural network model we call QuasiNet. Our model resembles the pi-sigma network \cite{ghosh1992}, except that we apply the hyperbolic tangent as an activation function on the hidden layer, thus obtaining hidden activations from the interval $(-1,1)$. We then multiply these on the output layer. The weights in our model are not fixed, nor do they act as exponents, but are involved in the computation so that at one extreme a given hidden layer activation does not matter (it is reduced to one as if the neuron activations were raised to 0) and at the other extreme it performs at its original function (unchanged as if they were raised to 1). This recalculation, which replaces the exponentiation from the previous models, is in fact polynomial. This property guarantees that complex numbers do not appear anywhere in the computation, and also allows the model to learn which hidden neurons ,,make sense'' to use to achieve the desired output. We show a schematic representation of the model in Fig.~\ref{fig:ourmodel}. For the ease of explaining the workings of the model we will describe a simple case with one summation and one product layer. However, QuasiNet can be extended to any number of connected layers and of various types (both product and summation) and still can be trained with the same standard error backpropagation learning rule \cite{rumelhart1986learning}.

In our current research, we chose the hyperbolic tangent because by representing the logical values truth and false by $-1$ and $+1$ we can directly simulate the logical XOR function using multiplication, since sign multiplication behaves exactly like XOR. For other tasks, it may be appropriate to choose a different activation function.

\subsection{Forward pass}

Given an input $\boldsymbol x$, including the trainable bias, and weights $w^{\rm hid}$ and $w^{\rm out}$ we compute the activation of neurons on the connected hidden layer $h$ as:
\begin{equation}
    h_i = \tanh ( \sum_{j} w^{\rm hid}_{ij} \cdot x_j ),
\end{equation}
and from that the activation of the connected product layer $y$, which in this case, is the output layer of the model, can be expressed as
\begin{equation}
    y_i = \prod_{j} \left( 1 - \sigma (w^{\rm out}_{ij}) ( 1 - h_j) \right),
\end{equation}
or
\begin{equation}
    y_i = \prod_{j} f( h_j, \sigma (w^{\rm out}_{ij})),
\end{equation}
where $f$ is defined as:
\begin{equation}
    f(h_j, d) =  1 - d ( 1 - h_j)
\end{equation}
and $\sigma$ is the logistic function.

Thus, our model uses the analogy of the product neuron \cite{durbin1989} at the output layer, but replaces the input-weight exponentiation by a polynomial function $f$ with one nonlinear parameter:
\begin{equation}
	f(h, \sigma(w)) = 1 - \sigma(w) (1 - h)
\end{equation}

This approach preserves some properties of the exponentiation, namely $h^1 = f(h,1) = h$ and $h^0 = f(h,0) = 1$ but does not preserve the property $0^d = 0$. 
Instead, in the case of zero input, we get $f(0, d) = 1-d$.

Moreover, our method of applying weights using the function $f$ is continuous and continuously differentiable for all zero input values, while $0^0$ is undefined and hence a point of discontinuity and without derivative. By applying the logistic function to the weight, we achieve that the weight can take arbitrary real values, but which correspond to the exponents of $0$ and $1$ at the extrema. This allows the output weights to learn safely using gradient methods which would otherwise put them outside of the desired values.
 
\subsection{Error backpropagation}

For the output layer, we can use the gradient descent method for the mean squared error to derive the following rule for adapting the hidden-output weights:
\begin{equation}
\begin{aligned}
    %\Delta w^{out}_{ij} \sim
    \frac {\partial E} {\partial w^{\rm out}_{ij}} = {} 
    & ( d_i - y_i ) \left( \prod_{k \neq j} 1 - \sigma(w^{\rm out}_{ik})( 1 - h_k) \right)\\
    & (h_j - 1) \sigma (w^{\rm out}_{ij}) (1 - \sigma (w^{\rm out}_{ij}))
\end{aligned}    
\end{equation}
and back-propagating the error to the hidden layer:
\begin{equation}
    \frac {\partial E} {\partial h_i} =
    \sum_{k} ( d_k - y_k ) \left( \prod_{l \neq i} 1 - \sigma(w^{\rm out}_{kl})( 1 - h_l) \right) \sigma (w^{\rm out}_{ki})
\end{equation}

To speed up the computation, we can replace the partial multiplication in the learning rule by division, in case division by zero does not occur in the computation:
\begin{equation}
    \prod_{k \neq j} 1 - \sigma(w^{\rm out}_{ik})( 1 - h_l) = \frac{y_i}{1-\sigma(w^{\rm out}_{ij})(1-h_j)}
\end{equation}

Then we used the back-propagated error to adapt the input-hidden weights as:
\begin{equation}
    \Delta w^{\rm hid}_{ij} \sim
    \frac {\partial E} {\partial w^{\rm hid}_{ij}} =
    \frac {\partial E} {\partial h_i} (1 - h_i^2) x_j
\end{equation}

Our QuasiNet model does not need to be further constrained or stabilized; it can be approached in all respects as a multi-layered perceptron. The multiplication layer can also be incorporated into the network in other ways, such as alternating the summation and multiplication layers, multiplying first and then adding, and so on. It is also possible to modify or rescale the $f$ function according to the properties we want to achieve and the model will look very similar.

\section{Experimental results}

\subsection{XOR and Parity}

For the baseline comparison, we used a basic multilayer perceptron (MLP) without any regularization.
We implemented our model in Python programming language\footnote{url{https://github.com/kik-re/prodnet}} by modifying the original MLP to make the comparison as faithful as possible.
Here we present mainly the results from experiments with the XOR and $n$-parity problem and preliminary results from other problems. In general, XOR is indeed a parity problem, but we keep the naming for convenience.

The parity problem, i.e., determining whether a binary number has an even or odd number of units, is a classical, rather hard problem for neural networks and is used to test and validate new models.
In our experiments, we evaluate the convergence, i.e., how many networks out of 100 reach a stable solution.
We consider that the network converged to the solution if it gives the correct output for ten consecutive epochs.
We gave networks at most 10 thousand training epochs, in the case of larger $n$-parity.
For both types of networks, we use a uniform learning rate $\alpha = 0.9$ and a Gaussian initialization of the weights with distribution $\mathcal{N}(0; 1.0)$.
These hyperparameters were chosen based on previous experimentation.

\begin{figure}[ht]
	\centering
	\vspace{1em}
 	\includegraphics[width=.49\textwidth]{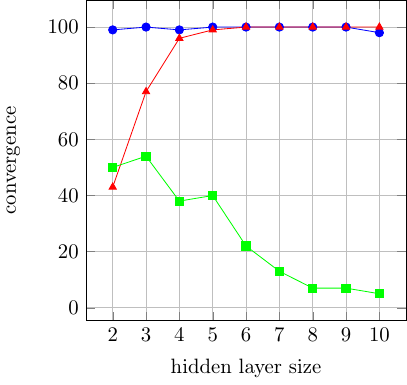}
	\includegraphics[width=.49\textwidth]{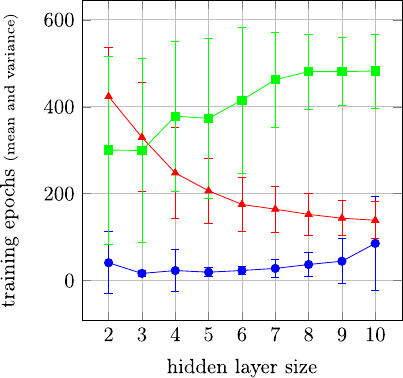}\vspace{0.5em}
	\includegraphics[width=.65\textwidth]{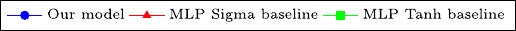}
	\caption{Results from XOR experiments with varying hidden layer size (max.~500 epochs) in terms of convergence (left) and average training epochs to convergence (right), including non-converging runs in the mean.}
	\label{fig:xor-hidsize}
\end{figure}

In Fig.~\ref{fig:xor-hidsize}, we show the effect of hidden layer size on the convergence of networks in the XOR task.
To emphasize the point that QuasiNet is not successful solely because of encoding the labels of the problem as $-1,1$ instead of $0,1$ we include the results for the baseline with two consecutive hyperbolic tangent layers.
For the so-called minimal XOR with 2 hidden neurons, we observe a clear superiority of our model.
So far, there is no three-layer network model known to us without residual weights that learns this problem to 100\% in the minimal architecture with 2 hidden neurons, i.e., that all trained networks find a solution.

In Table~\ref{tab:parity-results} we show the results of initial experiments for parity $2-7$, where parity 2 represents XOR. For MLP, we no longer found an optimal hidden layer size for parity greater than 5. Given the constraint on the number of training epochs we set above, MLP could not learn the task.

\begin{table}[ht]
	\centering
	\def\arraystretch{1.5}
	\begin{tabular}{|@{\hspace{2mm}}l@{\hspace{2mm}}|@{\hspace{2mm}}c@{\hspace{2mm}}|@{\hspace{2mm}}c@{\hspace{2mm}}|@{\hspace{2mm}}c@{\hspace{2mm}}|@{\hspace{2mm}}c@{\hspace{2mm}}|}
		\hline
		 & \multicolumn{2}{c|}{\bf QuasiNet} & \multicolumn{2}{@{\hspace{2mm}}c@{\hspace{2mm}}|}{\bf MLP baseline} \\
		\hline
		$n$-parity & {\bf h} & convergence & {\bf h} & convergence \\
		\hline
		{\bf 2} & 2 & 100 & 4 & 100 \\
		\hline
		{\bf 3} & 4 & 100 & 9 & 100 \\
		\hline
		{\bf 4} & 6 & 100 & 12 & 91 \\
		\hline
		{\bf 5} & 7 & 100 & 50 & 44 \\
		\hline
		{\bf 6} & 12 & 100 & 45 & 69 \\
		\hline
		{\bf 7} & 15 & 100 & 45 & 33 \\
		\hline
	\end{tabular}  
	\vspace{0.5em}
  \caption{Results from n-parity experiments: minimum size of the hidden layer {\bf h} for maximum number of converging networks. For MLP baseline we report results we have achieved given by the computational limits, very large hidden layer size would lead to a slightly better performance, but not full convergence.}
  \label{tab:parity-results}
\end{table}

\begin{figure}[ht]
	\centering
	\vspace{1em}
  	\includegraphics[width=.4\textwidth]{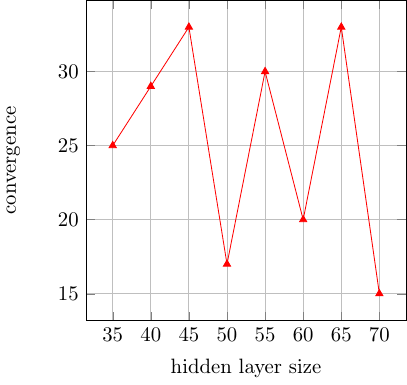}
 	\includegraphics[width=.4\textwidth]{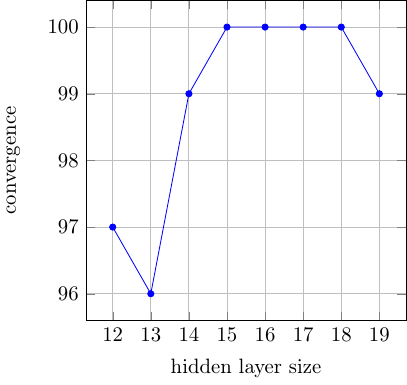}
  	\includegraphics[width=.4\textwidth]{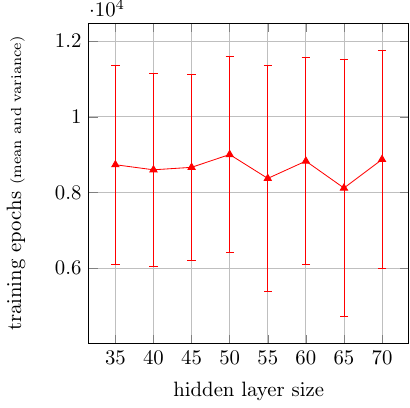}
 	\includegraphics[width=.4\textwidth]{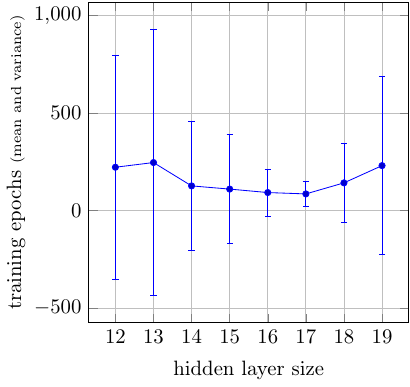} 	
	\caption{Results from parity 7 experiments: convergence (top), training epochs (bottom) for MLP baseline (left) and QuasiNet (blue).}
	\label{fig:parity7-hidsize-baseline}
\end{figure}

In figure \ref{fig:parity7-hidsize-baseline} we display the results of hidden layer size experiments for parity 7 for both our model and the MLP baseline.
Note, that we allowed the MLP baseline to train for 10 thousand epochs, which is significantly more then in the case of our QuasiNet, which in the best case converges in less then 100 epochs.
Similarly, the tested hidden layer sizes for the MLP baseline and its success differ dramatically from our model.

\begin{figure}[ht]
	\centering
	\vspace{1em}
 	\includegraphics[width=.45\textwidth]{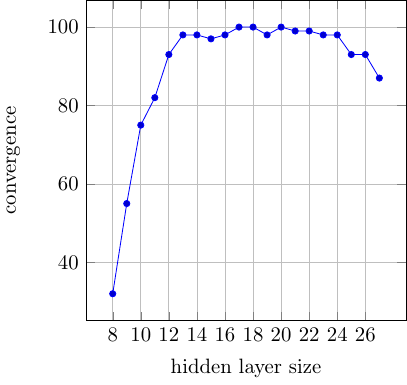}
	\includegraphics[width=.45\textwidth]{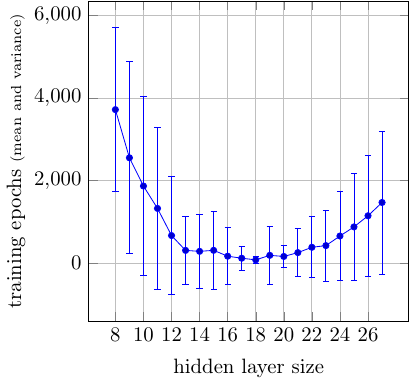}
	\caption{Results from parity 8 experiments with varying hidden layer size (100~nets, max.~5000 epochs) in terms of convergence (left) and average training epochs to convergence (right), including non-converging runs in the mean.}
	\label{fig:parity8-hidsize}
\end{figure}

In figure \ref{fig:parity8-hidsize} we display a more detailed experiment with QuasiNet: convergence in parity 8 as a function of the hidden layer size. 
We also show the mean training epochs. Note, that for deeper evaluation we allowed the model to train for 5000 instead of 500 epochs. 
This experiment illustrates that larger hidden layer size does not lead to better performance.
The optimal hidden layer size for out model seems to be rather close to the number of input features.

\begin{figure}[ht]
	\centering
	\vspace{1em}
 	\includegraphics[width=.9\textwidth]{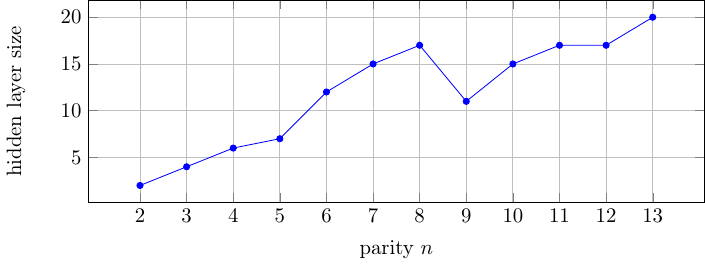}
	\caption{Results from $n$-parity experiments: minimum size of the hidden layer {\bf h} for maximum number of converging networks.}
	\label{fig:parity-all}
\end{figure}

Last, but not least, we present overal results for parity problem with in terms of best hidden layer size found so far for $2 \le n \le 13$ in Fig~\ref{fig:parity-all}. 
%For experiments with larger $n$ we change the initial weight distribution to $\mathcal{N}(0; 0.5)$ and maximum number of epochs to 1 thousand. 
In our experiments we observe that most of the nets converge in relative small number of epochs even for large $n$-parities.

\subsection{Multiple layers and 2 spirals}

\begin{figure}[htb]
	\centering
	\vspace{1em}
 	\includegraphics[width=\textwidth]{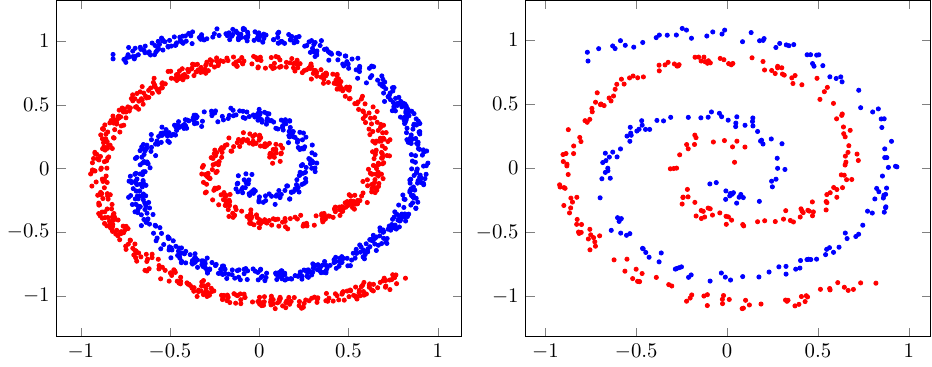}
	\caption{The 2 spirals dataset (2000 pts) split into 80\% training and 20\% testing data.}
	\label{fig:spirals-data}
\end{figure}

The famous 2 spiral problem is commonly used for testing new neural network models.
Similarly to parity it poses a problem with mutually exclusive situations,~i.e the point belongs to one spiral or to another.
We display the 2 spirals dataset with 2 thousand points we have used and its distribution into training and testing data in Fig.~\ref{fig:spirals-data}. 
Note, that the spiral coordinates are transformed to the interval $(-1,1)$ similarly to inputs in parity problem.
In our preliminary experiments we have observed that this problem requires more hidden layers for a satisfying performance level.

\begin{figure}[htb]
	\centering
	\vspace{1em}
 	\includegraphics[width=\textwidth]{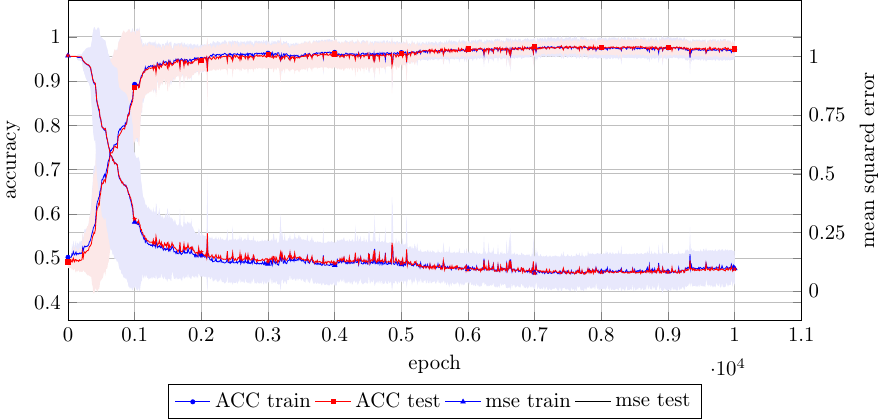}
	\caption{Results from training of the 2 spirals problem: mean and standard deviation over 10 nets trained for 10000 epochs, only every 10th epoch is shown.}
	\label{fig:spirals-training}
\end{figure}

In Fig~\ref{fig:spirals-training} we display training progress of QuasiNet with four hidden layers of neurons, where the hyperbolic tangent and product layers are combined one after another.
Namely, the architecture used was: 2 input neurons, 10 tanh neurons, 80 product neurons, 5 tanh neurons and finally 1 output product neuron.
Other hyperparameters used were learning rate $\alpha = 0.01$ and a Gaussian initialization of the weights with distribution $\mathcal{N}(0; 0.5)$.
The networks were trained for 10 thousand epochs.  
The mean accuracy for the testing data set was 98.275\% and 4 out of 10 networks achieved full convergence, which we deem very successful. 

\newpage
\section{Conclusion and future work}

Our innovative neural network model QuasiNet compensates for particular shortcomings of existing neural networks with product neurons and proposes a new way to train the weights of any multiplicative components in neural networks, which avoids the problem with computations in the domain of complex numbers. Our model is mathematically simple, easy to implement, and can be combined with existing models, with the promise of increasing their efficiency in tasks with mutually exclusive situations, which are in general quite difficult for classical neural networks to solve.
QuasiNet outperforms the classical MLP baseline in the XOR and $n$-parity tasks and excels also in the 2 spiral problem task, which is considered one of the hardest problems for neural networks in general.
A major advantage of QuasiNet is that it needs a quite small amount of parameters and training epochs to learn hard tasks.
From the preliminary queries in the literature and results from related models we assume QuasiNet is at least comparable with the existing baselines. 
Yet a thorough experimentation and comparison outside of the scope of this debut research paper is to be done to strengthen our claims.

There is a vital need to continue the experimentation with this intriguing new model in terms of testing of various classical neural networks problems and comparing it to existing work.
We believe that in the future our model could be also beneficial in the domain of deep learning where it could be used instead of classical fully connected layers in combination with convolutional layers for feature extraction. 
The fact that the model requires only a fraction of weights compared to the classical fully-connected MLP with Sigmoid or similar activation function could be utilized for analyzing the properties encoded in the feature space of deep networks thus contributing to better explainability of the model.
The properties of QuasiNet, its explainability and possibility to encode mutually exclusive properties would also benefit the neural modeling in the domain of cognitive robotics and human-robot interaction where the development of the understanding and interpretation of the sensory input is vital for the robotic system. 

\bigbreak 

\noindent
{\bf Acknowledgements} The research leading to these results has received funding from the project titled TERAIS in the framework of the program Horizon-Widera-2021 of the European Union under the Grant agreement no. 101079338.

\bibliographystyle{splncs04}
\bibliography{malinovska-etal-icann2023}

\end{document}